\documentclass{article}

\usepackage[preprint,nonatbib]{neurips_2023}

\usepackage[utf8]{inputenc} 
\usepackage[T1]{fontenc}    
\usepackage{hyperref}       
\usepackage{url}            
\usepackage{booktabs}       
\usepackage{amsfonts}       
\usepackage{nicefrac}       
\usepackage{microtype}      
\usepackage{xcolor}         
\usepackage{amsmath}
\usepackage{amssymb}
\usepackage{booktabs}

\usepackage[utf8]{inputenc}  
\usepackage[T1]{fontenc}
\usepackage[nonatbib]{neurips_2023}

\usepackage{makecell}
\usepackage{multirow}
\usepackage{caption}
\usepackage{subcaption}
\usepackage{cleveref}
\usepackage{mathtools}
\usepackage{csquotes}
\Crefname{figure}{Figure}{Figures}
\Crefname{section}{Section}{Sections}
\Crefname{equation}{Eq.}{Eqs.}
\Crefname{proposition}{Prop.}{Props.}
\Crefname{theorem}{Thm.}{Thms}
\Crefname{lemma}{Lemma}{Lemmas}
\Crefname{definition}{Def.}{Defs}
\Crefname{algorithm}{Alg.}{Algs}
\Crefname{remark}{Remark}{Remarks}

\usepackage{tikz,pgf}
\usetikzlibrary{positioning,shapes,shapes.misc,arrows,calc,arrows.meta,fit, backgrounds,quotes,intersections,patterns,decorations.pathmorphing,decorations.pathreplacing,decorations.markings}
\tikzset{
  >={Latex[width=1.5mm,length=1.7mm]},
  font=\sffamily\scriptsize,   
  RR/.style={draw,circle,inner sep=0mm, minimum size=4.mm,font=\sffamily\scriptsize},
  intervened/.style={RR,draw=red,red,thin},
  AUX/.style={draw=none,inner sep=0mm,font=\sffamily\scriptsize},
  unobserved/.style={RR,fill=black!40,text=white,draw=black!40},
  every picture/.style=semithick,
  snakey/.style={decorate, decoration={snake,segment length=2mm, amplitude=.25mm,pre length=.5mm, post length=.5mm}},
  given/.style={fill=gray!50}
}

\title{Can We Utilize Pre-trained Language Models within Causal Discovery Algorithms?}
\author{%
    Chanhui Lee$^{1}$\thanks{These authors are contributed equally.} , Juhyeon Kim$^{2*}$, Yongjun Jeong$^3$, Juhyun Lyu$^4$ , Junghee Kim$^4$, \\
    \textbf{Sangmin Lee}$^4$, \textbf{Sangjun Han}$^4$, \textbf{Hyeokjun Choe}$^4$,  \textbf{Soyeon Park}$^4$, \textbf{Woohyung Lim}$^4$, \\ \textbf{Sungbin Lim}$^{5, 6}$\thanks{Corresponding Authors.} , \textbf{Sanghack Lee}$^{2,7\dagger}$\\[.35em]
    $^1$Department of Artificial Intelligence, Korea University \\ $^2$Graduate School of Data Science, Seoul National University \\ $^3$Department of Computer Science and Engineering, UNIST \\ $^4$Data Intelligence Laboratory, LG AI Research \\ $^5$Department of Statistics, Korea University \\ $^6$LG AI Research \\ $^7$SNU-LG AI Research Center
}

\begin{document}
\maketitle
\begin{abstract}
Scaling laws have allowed Pre-trained Language Models (PLMs) into the field of causal reasoning. 
Causal reasoning of PLM relies solely on text-based descriptions, in contrast to causal discovery which aims to determine the causal relationships between variables utilizing data.
Recently, there has been current research regarding a method that mimics causal discovery by aggregating the outcomes of repetitive causal reasoning, achieved through specifically designed prompts \cite{kichiman2023}.
It highlights the usefulness of PLMs in discovering cause and effect, which is often limited by a lack of data, especially when dealing with multiple variables. 
Conversely, the characteristics of PLMs which are that PLMs do not analyze data and they are highly dependent on prompt design leads to a crucial limitation for directly using PLMs in causal discovery.
Accordingly, PLM-based causal reasoning deeply depends on the prompt design and carries out the risk of overconfidence and false predictions in determining causal relationships.
In this paper, we empirically demonstrate the aforementioned limitations of PLM-based causal reasoning through experiments on physics-inspired synthetic data.
Then, we propose a new framework that integrates prior knowledge obtained from PLM with a causal discovery algorithm. 
This is accomplished by initializing an adjacency matrix for causal discovery and incorporating regularization using prior knowledge.
Our proposed framework not only demonstrates improved performance through the integration of PLM and causal discovery but also suggests how to leverage PLM-extracted prior knowledge with existing causal discovery algorithms.
\end{abstract}

\section{Introduction}
\label{introduction}
Discovering causal structure, which is often represented as a directed acyclic graph (DAG) is a crucial problem across diverse scientific and industrial fields \cite{de2004discovery, addo2021exploring}. 
The fact that the number of possible DAGs grows super exponentially as the number of variables increases 
can lead to a data scarcity problem, which is a potential limitation when discovering a causal structure among a large number of variables.
One approach to address the issue is using prior knowledge \cite{Borboudakis2011ACA, Borboudakis2012IncorporatingCP, Kalainathan2018SAMSA, sinha2021perturbing}. 
For example, one can determine a graph from a Markov equivalence class by figuring out edge direction using prior knowledge of specific edges \cite{Borboudakis2012IncorporatingCP, sinha2021perturbing}

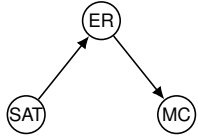
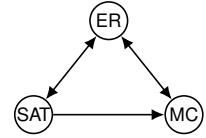
\begin{figure}[t]
    \centering
    \begin{subfigure}[b]{.375\columnwidth}\scriptsize
    Which of the following causal relationship is correct?\\[.25em]
A. Changing $\{\alpha\}$ can directly change $\{\beta\}$.\\
B. Changing $\{\beta\}$ can directly change $\{\alpha\}$.\\
C. Both A and B are true.\\
D. None of the above. No direct relationship exists.\\[.25em]
Let’s think step-by-step to make sure that we have the
right answer.
Then provide your final answer within the tags,
$\langle$Answer$\rangle$ A/B/C/D $\langle$/Answer$\rangle$
    \caption{prompt template}
    \label{fig:causalreasoning-a}
    \end{subfigure}\hfill\begin{subfigure}[b]{.3\columnwidth}\centering
    \begin{tikzpicture}[x=2cm,y=2.5cm]
        \node[RR,minimum size=5mm] (ER) at (0.5,0.5) {ER};
        \node[RR,minimum size=5mm] (SAT) at (0,0) {SAT};
        \node[RR,minimum size=5mm] (MC) at (1,0) {MC};
        \draw[->] (SAT) -- (ER);
        \draw[->] (ER) -- (MC);
    \end{tikzpicture}
    \caption{ground truth}
    \label{fig:causalreasoning-b}
    \end{subfigure}\hfill
    \begin{subfigure}[b]{.3\columnwidth}\centering
    \begin{tikzpicture}[x=2cm,y=2.5cm]
        \node[RR,minimum size=5mm] (ER) at (0.5,0.5) {ER};
        \node[RR,minimum size=5mm] (SAT) at (0,0) {SAT};
        \node[RR,minimum size=5mm] (MC) at (1,0) {MC};
        \draw[<->] (SAT) -- (ER);
        \draw[<->] (ER) -- (MC);
        \draw[->] (SAT) -- (MC);
    \end{tikzpicture}
    \caption{PLM prediction}
    \label{fig:causalreasoning-c}
    \end{subfigure}
    \caption{
    (a) A multiple-choice template for determining causality. 
    (b) Ground truth and (c) PLM prediction for a causal graph consisting of Surface Air Temperature (SAT), Evaporation Rate (ER), and the Moisture Content of the object (MC) using the prompt template.
    }
    \label{fig:causalreasoning}
\end{figure}

Recent breakthroughs in Pre-trained Language Models (PLMs) \cite{Wei2022EmergentAO, openai2023gpt4, anil2023palm, touvron2023llama} 
have demonstrated its potential for diverse reasoning tasks.
Given the broad spectrum of text corpora utilized during pre-training, it is known that PLMs can, by employing specifically crafted task descriptions known as prompts, perform a wide range of real-world tasks including commonsense and numerical reasoning \cite{Suzgun2022ChallengingBT}, code generation \cite{Chen2021EvaluatingLL}, and dialogue generation \cite{Thoppilan2022LaMDALM}.
The growing reasoning capability of PLM enables reasoning-based causal discovery via a designed prompt template (see Figure \ref{fig:causalreasoning-a}), which uses variable names as nodes for a causal graph
\cite{choi2022lmpriors, kichiman2023}.
In particular, by employing chain-of-thought prompting \cite{wei2023chainofthought}, which decomposes a problem into a series of simpler tasks, Kıcıman et al. exhibited the potential of PLM in causal discovery by outperforming conventional causal discovery algorithms on some benchmark datasets \cite{kichiman2023}.
Besides the potential in performance, relying only on variable names makes PLM-based causal reasoning bypass the data scarcity problem. 

However, 
this approach has inherent limitations with respect to the following aspects. 
First, as PLM-based causal reasoning relies on variable names alone, the discovered causal graph does not reflect rich information in underlying data.
Second, the black-box nature of PLMs means that we can not inspect the process of causal reasoning.
The third limitation, which we will exemplify soon, is the absence of a prompt design that allows PLMs to perceive entire variables, failing to distinguish direct and indirect causal relationships.
Therefore we suggest our framework which integrates causal reasoning of PLM and data-driven causal discovery to overcome those limitations.

\paragraph{Motivation}
We provide a motivating example based on physics to explain the potential of the integration of data-driven causal discovery and PLM-based causal reasoning to overcome the limitations. 
First, we illustrate the limitations through a causal reasoning example based on GPT-4 in \Cref{fig:causalreasoning} where \Cref{fig:causalreasoning-a} shows the prompt modified from that of \cite{kichiman2023}.
\Cref{fig:causalreasoning-b} depicts the ground truth where SAT, ER, and MC, respectively, represents surface air temperature, evaporation rate, and moisture content of the object.
\Cref{fig:causalreasoning-c} shows the case in which PLM would falsely predict indirect relation as if direct relation because in the prompt, only pairs of variables are examined without information of other nodes. 
Furthermore, 
the resulting graph contains bidirectional edges\footnote{In some literature, bidirected edges are employed to represent unmeasured confounding. In this work, a bidirected edge represents two directed edges indicating both causal relations hold true.}, violating acyclicity.

\begin{figure}[t]
\begin{subfigure}{.225\columnwidth}\centering
\begin{tikzpicture}[x=1.8cm,y=1.5cm]
\node[RR,minimum size=5mm] (AP) {AP};
\node[RR,minimum size=5mm] (TSI) at (0.5, 0.5) {TSI};
\node[RR,minimum size=5mm] (ER) at (1,0) {ER};
\draw[->] (AP) -- (TSI);
\draw[->] (TSI) -- (ER);
\end{tikzpicture}
\caption{ground truth}
\label{fig:virtual_graph-a}
\end{subfigure}\hfill%
\begin{subfigure}{.225\columnwidth}\centering
\begin{tikzpicture}[x=1.8cm,y=1.5cm]
\node[RR,minimum size=5mm] (AP) {AP};
\node[RR,minimum size=5mm] (TSI) at (0.5, 0.5) {TSI};
\node[RR,minimum size=5mm] (ER) at (1,0) {ER};
\draw[->] (TSI) -- (ER);
\end{tikzpicture}
\caption{PLM}
\label{fig:virtual_graph-b}
\end{subfigure}\hfill\begin{subfigure}{.225\columnwidth}\centering
\begin{tikzpicture}[x=1.8cm,y=1.5cm]
\node[RR,minimum size=5mm] (AP) {AP};
\node[RR,minimum size=5mm] (TSI) at (0.5, 0.5) {TSI};
\node[RR,minimum size=5mm] (ER) at (1,0) {ER};
\draw[->] (AP) -- (TSI);
\draw[->] (TSI) -- (ER);
\draw[->] (AP) -- (ER);
\end{tikzpicture}
\caption{DAG-GNN}
\label{fig:virtual_graph-c}
\end{subfigure}\hfill%
\begin{subfigure}{.225\columnwidth}\centering
\begin{tikzpicture}[x=1.8cm,y=1.5cm]
\node[RR,minimum size=5mm] (AP) {AP};
\node[RR,minimum size=5mm] (TSI) at (0.5, 0.5) {TSI};
\node[RR,minimum size=5mm] (ER) at (1,0) {ER};
\draw[->] (AP) -- (TSI);
\draw[->] (TSI) -- (ER);
\end{tikzpicture}
\caption{our method}
\label{fig:virtual_graph-d}
\end{subfigure}    
    \caption{
    The combination of relationships that are challenging to capture due to unknown associations and data noise.
    (a) Ground truth causal relation of the previously unobserved variable for PLM (Apollon, AP) and previously observed Total Solar Irradiance (TSI) and Evaporation Rate (ER), and (b, c, d) the predicted causal relations by PLM, DAG-GNN \cite{yu2019daggnn}, and our approach.
    }
    \label{fig:virtual_graph}
\end{figure}
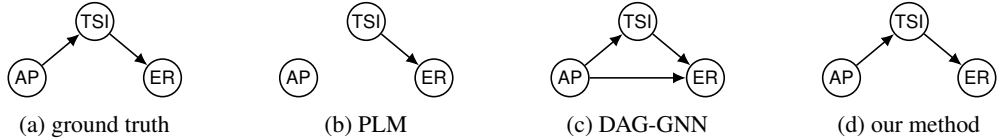

Consider a chain causal structure (\Cref{fig:virtual_graph-a}), introduced in \Cref{fig:causalreasoning} where an imaginary variable \textbf{Apollon} (AP) is inserted for illustrative purposes.
The causal relationship between Total Solar Irradiance (TSI) and its impact on the evaporation rate (ER) exhibits a weak connection, primarily attributed to significant noise within the data; nonetheless, it still reflects a well-established fact. 
On the other hand, a different edge between Apollon (AP) and TSI symbolizes a newly emergent causal relation, which the PLM cannot discern, perhaps due to the object being discovered after the PLM's pre-training or highly domain-specific.


For demonstration, we synthesized a dataset where Gaussian noise is scaled up between TSI and ER than that between AP and TSI so that the cause-effect relation of the noisy edge would still be clear for a PLM but less clear for typical causal discovery algorithms to detect (details in \Cref{supple:generation}).
In contrast, the cause-effect relation of the imaginary edge would be easily identifiable for causal discovery algorithms but not for a PLM.
Indeed, according to our experimental results, the PLM easily determines the causal relation of the noisy edge but, as anticipated, struggles to recognize the causal relation relating to the novel variables (\Cref{fig:virtual_graph-b}).
On the other hand, the causal discovery algorithm (\Cref{fig:virtual_graph-c}) identifies direct causation for the imaginary edge. 
However, due to the significant noise, the causal discovery algorithm discovers false causal relations between AP and ER.


\paragraph{Contributions}
The above examples naturally raise the question: \begin{displayquote}Can we utilize a PLM with a causal discovery algorithm, taking advantage of both methods?\end{displayquote}
To this end, we propose a framework to pioneer the potential of integration of causal discovery and PLMs. 
Our contributions are summarized as follows.
\begin{itemize}
    \item 
    We reveal the limitations of PLM-based causal reasoning, such as non-adherence to DAG constraint and false discovery due to limited prompt design. 
    To investigate the limitations, we propose a method to construct a synthetic dataset, based on physical commonsense knowledge for evaluating our framework.
    \item 
    We propose a PLM-integrated causal discovery framework that combines PLM-generated graphs as prior knowledge in causal discovery algorithms.
    The proposed framework represents a convergence of data-driven and knowledge-driven methodologies, opening up an option for improved causal inference.
    \item 
    We show that the proposed framework could generally improve the performance of existing causal discovery algorithms across synthetic and real datasets.
\end{itemize}

\section{Preliminaries}
\label{sec:preliminaries}

Causal discovery seeks to identify a causal graph that represents causal relations among variables from data \cite{Chickering2003OptimalSI,spirtes2016causal, spirtes2000causation,GlymourZhangSpirtes2019_reviewb}.
Among many approaches to causal discovery, a score-based approach uses a score function to evaluate candidate graphs based on how well a graph explains observed data. 
In this section, we briefly introduce three score-based methods used in this paper.

Non-combinatorial Optimization via Trace Exponential and Augmented Lagrangian for Structure Learning (NOTEARS) \cite {zheng2018notears} proposed a DAG constraint using matrix exponential so as to change a combinatorial optimization problem over DAGs into a continuous optimization problem. 
Given $d$ variables, a causal graph can be expressed as a structural coefficients matrix $\mathbf{W}\in \mathbb{R}^{d \times d}$ under a linear assumption.
Given a dataset $\mathbf{X}\in\mathbb{R}^{n \times d}$ with $n$ observation, NOTEARS uses a training objective as follows.
\begin{align}
\underset{\mathbf{W} \in \mathbb{R}^{d \times d}}{\text{min }} &\quad L(\mathbf{W}):= \underbrace{\frac{1}{2n} \| \mathbf{X} - \mathbf{X}\mathbf{W} \|_{F}^{2}}_\text{Fitting Loss} + \underbrace{\lambda\| \mathbf{W} \|_{1}}_\text{Sparsity Loss}\\
\text{subject to}& \quad \operatorname{tr}(e^{\mathbf{W} \circ \mathbf{W}})-d=0. \notag
\end{align}
The loss is designed to capture underlying causal relations with a smaller number of edges (controlled by a hyperparameter $\lambda$) while ensuring the acyclicity of the learned graph.
A fitting loss aims to minimize the Frobenius norm between dataset $\mathbf{X}$ and coefficient matrix $\mathbf{W}$.
This is complemented by a sparsity loss, which helps reduce the occurrence of false discoveries where $\lambda$ is a hyperparameter for scaling sparsity loss.
In the end, a threshold sets the element of the structural coefficient matrix to 0 or 1, transforming it into an adjacency matrix, with weights below the threshold being rounded down and those above being rounded up.

We also employed DAG-GNN \cite{yu2019daggnn} and CGNN \cite{goudet2018learning}. 
DAG-GNN is a continuous optimization-based causal discovery algorithm. 
By explicit parameterization and the acyclicity constraint as NOTEARS, DAG-GNN directly learns a DAG's structural coefficient matrix.
Given $X\in\mathbb{R}^d, \mathbf{W} \in \mathbb{R}^{d \times d}$ representing a sample of $d$ variables and the corresponding structural coefficient matrix, DAG-GNN use the linear Structural Equation Modeling (SEM) assumption:
\begin{align}
\label{eq:linearsem}
    X &= \mathbf{W}^TX + Z
\end{align}
where each $Z\in \mathbb{R}^d$ is a latent variable representing random noise.
\Cref{eq:linearsem} can be modeled as 
\begin{equation}
\label{eq:daggnn_enc}
    Z = (I-\mathbf{W}^T)X
\end{equation}
\begin{equation}
\label{eq:daggnn_dec}
    X = (I - \mathbf{W}^T)^{-1}Z,
\end{equation}
which enables separated parametric modeling of \Cref{eq:daggnn_enc} as latent variable encoder and \Cref{eq:daggnn_dec} as decoder for the given latent variable.
With the encoder-decoder architecture, DAG-GNN is formulated as a variational autoencoder (VAE) \cite{kingma2013vae}, minimizing the evidence lower bound (ELBO).
Causal Generative Neural Network (CGNN) is a differentiable generative model that first constructs a skeleton graph and then refines the skeleton graph via a greedy procedure.
Defining score function for given data, CGNN optimizes the skeleton graph by reverse, adding, or removing edges.

\section{Causal Discovery with PLM-derived Priors}
\label{method}

\begin{figure*}[t]%
    \centering%
    \includegraphics[clip,trim=10mm 20mm 10mm 15mm,width=\textwidth]{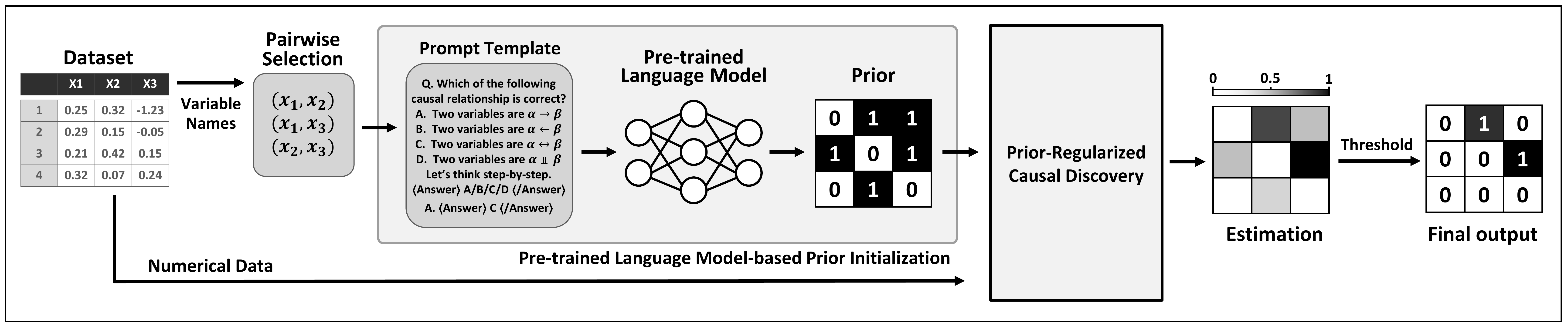}%
    \caption{
    Overview of the proposed framework. 
    For a dataset composed of observation of variables and corresponding names, PLM-based causal reasoning predicts an adjacency matrix as prior, using only the names of variables.
    With the prior extracted and observations in the dataset, train a causal discovery algorithm with the support of the prior extracted for adjacency matrix initialization and regularization.
    }
    \label{fig:framework}
\end{figure*}

The following sections illustrate methods for incorporating PLM prior knowledge $\mathbf{K} \in \{0, 1\}^{ d \times d}$ with optimization-based causal discovery algorithms.
The overall framework is depicted in \Cref{fig:framework}.
Given a dataset containing $d$ variables and the corresponding names, select pairs of variable names to complete the prompt template.
PLM proceeds causal reasoning given the prompts for each pairwise variable to complete PLM prior $\mathbf{K}$.
Next, the causal discovery algorithm leverages data and the PLM prior to predict continuous values of the adjacency matrix.
One method for incorporating $\mathbf{K}$ is to initialize the initial graph for causal discovery algorithms using $\mathbf{K}$ instead of a simple, empty skeleton graph. 
Another method is to regularize the training of $\mathbf{W}$ with regard to $\mathbf{K}$.
The final prediction is completed through the subsequent thresholding step applied for post-processing the coefficient matrix.

\subsection{Graph Initialization via Prior Knowledge}

Many optimization-driven causal discovery methods, including NOTEARS and DAG-GNN, use structural coefficient $\mathbf{W} \in \mathbb{R}^{d \times d}$ to represent the adjacency matrix of the ground truth causal graph.
Typically, $\mathbf{W}$ is initialized as a zero matrix. 
However, we hypothesize that such an initialization for the structural coefficient matrix could lead to a suboptimal state by getting caught in local optima.

Given that $\mathbf{K}$ represents the adjacency matrix of a causal graph and the element of $\mathbf{W}$ is set between 0 and 1, we suggest initializing $\mathbf{W}=\lambda_\text{init} \mathbf{K}$, expecting that $\mathbf{K}$ of appropriate quality prevents $\mathbf{W}$ getting caught in local optima.
Here, the scaling factor $\lambda_\text{init}$ is introduced for adjustment of the large elementwise gaps between $\mathbf{K}$ and $\mathbf{W}$.

On the other hand, causal discovery algorithms that do not use a structural coefficient matrix can utilize $\mathbf{K}$ in different ways.
An example of utilizing $\mathbf{K}$ involves employing element-wise expectations on $\mathbf{K}$.
Given a PLM $f$ and input prompt $T_{i,j}$, we can think of PLM as a stochastic function that maps $f: T \to K$.
Then $\mathbb{E}_{K_{i,j}\sim f(T_{i, j})}[K_{i,j} ]$ so it changes $\mathbf{K} $ into $\mathbf{K_{\text{mean}}} \in \mathbb{R}^{d \times d}$.
By introducing the concept of probabilistic edges through $\mathbf{K_{\text{mean}}}$, we can explore a continuous range of edges beyond simple binary 0 or 1 values.
In particular, we select CGNN \cite{goudet2018learning} as a representative method to showcase the effectiveness of graph initialization, considering that CGNN modifies the skeleton graph by adding, removing, and reversing individual edges.
For this, we employed $\mathbf{K}_{\text{mean}}$ instead of $\mathbf{K}$ in CGNN to prevent CGNN being captured in local minimum originated from the discrete value of $\mathbf{W}$.
In addition, since creating the skeleton graph through GNN can be time-consuming, utilizing PLM-based prior information bypasses the time for making a skeleton graph.



\subsection{Regularization with Prior Knowledge}
We suggest incorporating an additional loss term for the regularization of $\mathbf{W}$ based on prior knowledge $\mathbf{K}$. 
The regularization term is designed to allow the causal discovery algorithms to learn the causal relationships from $\mathbf{K}$ that cannot be inferred solely through the data. 
This additive approach facilitates a model-agnostic utilization of $\mathbf{K}$ for causal discovery algorithms that employ structural coefficients matrix.

The proposed regularization loss minimizes $\text{$\ell_{1}$-regularization}$ between the prior and estimated adjacency matrix, with the definition below.

\begin{equation}  
    L_{\text{sim}}(\mathbf{W}) := \sum_{i,j} |(\sigma (t \vert W_{i,j} \vert) - {K}_{i,j})|
\label{eq:sim}
\end{equation}
Subscript $i, j$ means an edge has a direction from variable $i$ to variable $j$ in an adjacency matrix.
The hyperparameter $t$ adjusts the steepness of the sigmoid function.
Then, our goal is to find an optimal matrix $\mathbf{W}_{\ast}$ which satisfies
\begin{equation}
\mathbf{W}_{\ast} = \underset{\mathbf{W}}{\text{argmin }} L(\mathbf{W})  + \lambda L_\text{sim}(\mathbf{W}),
\end{equation}
where $\lambda$ is the hyperparameter for scaling prior loss.
Since $W_{i,j}$ can take negative values, only the magnitude of $\mathbf{W}$ is utilized. 
To align $\mathbf{W}$ closely with a given prior, an activation function $\sigma$ such as the sigmoid function is employed to guarantee the edge-ness of $W_{i,j}$, by mapping $W_{i,j}$ to $[0, 1]$.
Subsequently, the difference between the two edges is computed to calculate the $\ell_{1}$ norm value.
This approach allows for flexible application when utilizing optimization-based causal discovery methods with the structural coefficient matrix.


\section{Experiments}
\label{sec:experiments}
In the subsequent section, we describe our experiments and the performance of each algorithm conducted by the proposed framework for a physical commonsense-based synthetic dataset and two real-world datasets: Arctic Sea Ice \cite{huang2021arctic} and cellular signaling networks on protein expression \cite{sachs2005causal}, called Sachs. 
Experiments on the synthetic dataset highlight the limitations of PLM-based causal reasoning. 
Experiments on the Arctic Sea Ice dataset demonstrate the effectiveness of PLM-causal reasoning in data-scarce cases, while the Sachs dataset is employed to represent the efficacy under data-rich scenarios. 

\subsection{Experimental Setting}
We explain the details of the datasets, metrics, and experimental setup used.

\begin{figure}
    \centering
    \scalebox{.9}{\begin{tikzpicture}[x=3cm,y=3cm]
\node[RR,minimum size=8mm] (TSI) at (0, 0.75) {TSI};
\node[RR,minimum size=8mm] (RNFL) at (0.5, 1) {RNFL};
\node[RR,minimum size=8mm] (Wgt) at (1,0.75) {Wgt};
\node[RR,minimum size=8mm] (SAT) at (0,0.25) {SAT};
\node[RR,minimum size=8mm] (ER) at (0.5,0.5) {ER};
\node[RR,minimum size=8mm] (WS) at (0.5,0) {WS};
\node[RR,minimum size=8mm] (MC) at (1,0.25) {MC};
\draw[->] (TSI) -- (SAT);
\draw[->] (TSI) -- (ER);
\draw[->] (TSI) -- (WS);
\draw[->] (SAT) -- (ER);
\draw[->] (WS) -- (SAT);
\draw[->] (WS) -- (ER);
\draw[->] (ER) -- (RNFL);
\draw[->] (ER) -- (MC);
\draw[->] (RNFL) -- (MC);
\draw[->] (MC) -- (Wgt);
\end{tikzpicture}
}
    \caption{Physical knowledge-based synthetic graph with size 7. The components of the graph are Rainfall (RNFL), Total Solar Irradiance (TSI), Surface Air Temperature (SAT), Wind Speed (WS), Evaporation Rate (ER), Moisture Content of object (MC), and Weight of object (Wgt). }
    \label{fig:physics_graph}
\end{figure}
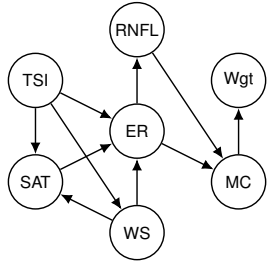

\paragraph{Physical commonsense-based synthetic dataset}


Since data often comes without the ground truth causal structure, the use of synthetic data with known causal structure is crucial to evaluate causal discovery methods.
In generating a synthetic dataset for evaluating PLM's causal reasoning and the proposed framework, it is necessary that the causal relationships among the variables in the graph are faithfully mirrored in real-world linguistic usage. 
This ensures PLM's ability to accurately infer a causal structure.
The condition is especially challenging in that causality embedded in the text is often in the realm of controversy.
%

We selected the physical commonsense domain
upon which most people can readily agree
given its long-standing history of experimental demonstrations in physics.
We collected event descriptions expressing the physical commonsense related to the evaporation of water, utilizing datasets including \cite{bisk2019piqa, Sap2019ATOMICAA}.
From the collected description of nodes and corresponding causal relations, we make a graph of different sizes, 7, 5, and 3 (see \Cref{fig:physics_graph} for the 7-node graph). To construct 5-node graph, we removed the Wind Speed and Weight of objects from the 7-node graph. Further, Rainfall and Total Solar Irradiance are removed from the 5-node graph to construct 3-node graph.
To generate data for each graph, 
we assumed linear relationships and independent Gaussian noises (see \Cref{supple:generation} for details).

\paragraph{Real-world datasets}

The \textbf{Arctic Sea Ice} dataset \cite{huang2021arctic} comprises 12 Earth science-related variables and only 486 instances. Its evaluation causal graph, constructed by a meta-analysis of literature referred in \cite{huang2021arctic}, contains 48 edges, including some bidirected edges and cycles. 
This dataset presents two challenges for conventional causal discovery algorithms due to 1) a small sample size and 2) possible discrepancies between the causal relationships in the underlying data and the ground truth.
Regardless of these challenges, PLMs are not affected since each causal relation in the ground truth is based on published papers, thus, PLM could have learned related knowledge.

The \textbf{Sachs} dataset \cite{sachs2005causal} consists of protein signaling pathways and comprises 11 variables with 7,466 observations. 
Its associated causal graph has a DAG structure with 19 edges \cite{ramsey2018fask}. 
The Sachs dataset, in contrast to the Arctic Sea Ice dataset, is a wealth of data and exhibits strong alignment with the causal graph. 
Given that the dataset uses abbreviations for variable names, which are highly domain-specific, we replaced these abbreviations with their full names to fully utilize PLM.


\paragraph{Metrics}

We report various metrics to evaluate the performance of each method from diverse perspectives.
SHD (Structural Hamming Distance) is the sum of the number of missing, extra, and reversed edges \cite{tsamardinos2006max} in the estimated graph (i.e., adjacency matrix). NHD and NHD ratio are the variants of SHD where NHD is SHD divided by the size of the matrix, and NHD ratio divides NHD by the worst case NHD.
Further, False Discovery Rate (FDR), False Positive Rate (FPR), and True Positive Rate (TPR) are employed to discern Type-I or Type-II errors \cite{li2009controlling} in causal discovery. 

\paragraph{Experimental setup}

We detailed baseline performances, hyperparameters, and settings for each algorithm. 
First, we employed GPT-4 as PLM since GPT-4 is known for its unique reasoning ability.
In the Arctic Sea Ice dataset, we reproduced the GPT-4 performance of \cite{kichiman2023} and used it for baseline.
We collected twenty GPT-4 causal reasoning results from the Sachs dataset and we selected PLM prior that reflects an average performance level over various metrics (40\%-65\% performance for FDR, TPR, FPR, SHD, NHD, and NHR ratio among 20 different PLM results).
We also replicated baseline performances of causal discovery to align closely with the results reported by \cite{zheng2018notears, yu2019daggnn, kichiman2023}.

Second, the hyperparameters of each algorithm are as follows:
The hyperparameter $t$ of NOTEARS is 10 and $\lambda$ of sparsity loss is 1 and $\lambda$ of prior similarity loss is 0.7 by the hyperparameter tuning.
We used the Adam optimizer for DAG-GNN and the architecture of the encoder and decoder in DAG-GNN consisted of two layers each. 
In terms of hidden features in the encoder and decoder, we allocated 64 features for the Arctic Sea Ice and 128 features for the Sachs.
CGNN does not use prior regularization in contrast to NOTEARS and DAG-GNN.
The reason is that CGNN does not use explicit modeling of the structural coefficient matrix, which is essential in prior regularization.


The \Cref{supple:experimental setup} provides metric definitions and offers details of hyperparameters and thresholds.

\begin{figure}[t]
    \centering
    \includegraphics[width=0.8\columnwidth]{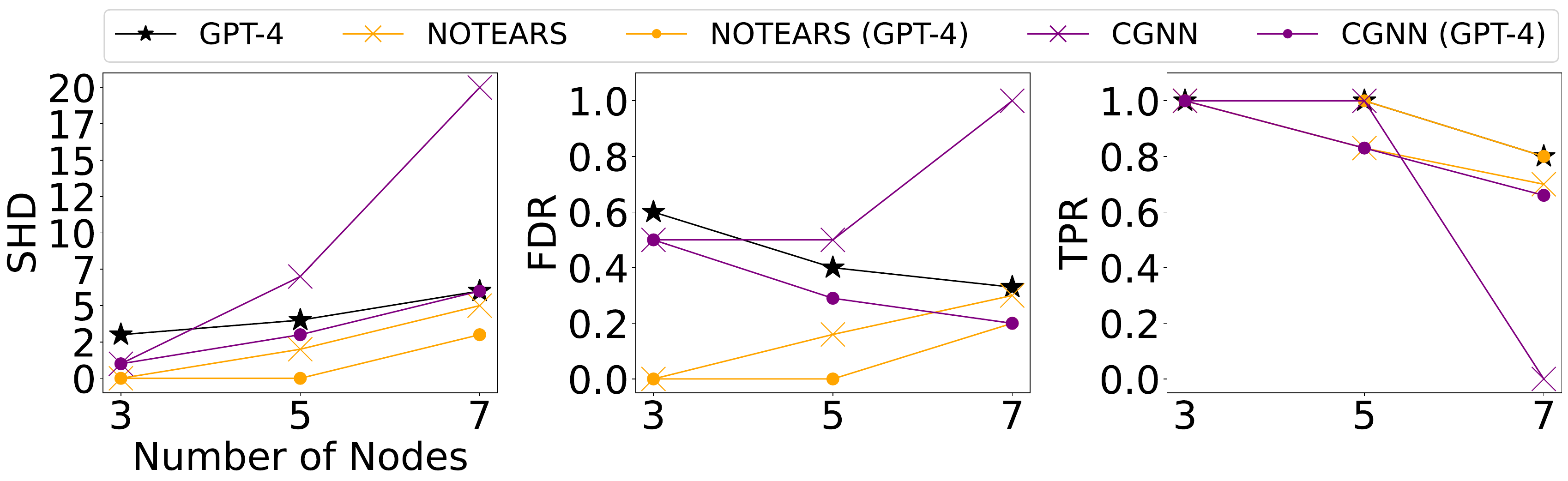}
    \caption{
    SHD, FDR, and TPR of NOTEARS and CGNN on the physical knowledge-based synthetic datasets with and without PLM prior.
    }
    \label{fig:synthetic results}
\end{figure}

\subsection{Empirical Results}
This section will report the experimental results and provide an analysis. 
Further, we explain our framework's noteworthy features and implications.

\paragraph{Physical synthetic dataset} 
We report in \Cref{fig:synthetic results} the SHD, FDR, and TPR for synthetic datasets. 
Overall, we observed that the integration of PLM prior improves performance when the number of nodes is larger than three (except for TPR of CGNN on five node dataset).
When the number of nodes is three, the causal graph of the dataset is too simple for NOTEARS so that it exactly predicted causal graphs of the dataset, resulting in no difference whether integrating PLM prior or not.
If the number of nodes is larger than three, vanilla NOTEARS fails to predict the causal graph, and integration of PLM prior brings out consistent performance enhancement for all metrics.

Similarly to NOTEARS, when the node size is smallest, CGNN showed no difference following the integration of PLM prior.
However, except for TPR, CGNN performance is improved with a huge difference, more than that of NOTEARS.
From the insights of \cite{goudet2018learning}, which indicate that utilizing priors closer to the ground truth graph enhances the performance of CGNN, we interpret that the use of PLM prior provides a promising skeleton graph.

Generally, the bigger the number of nodes gets, the harder the combinatorial problems are so SHD and TPR are getting worse.
In contrast, our framework mitigated the decline in performance than conventional causal discovery algorithms and GPT-4. 
NOTEARS is another representative causal discovery algorithm with which integration of PLM prior could enhance causal discovery performance. 
For the five and seven nodes datasets, NOTEARS shows enhancement of all the metrics concretely when integrated with PLM prior.

\paragraph{Arctic Sea Ice}

We present our findings in \Cref{tab:arctic_nhd}.
To start, it's important to note that Arctic Sea Ice has a limitation, as the causal graph is annotated based on a literature review, without a comprehensive examination of alignment among the sources.
This implies that the annotated causal graph could be misaligned with the ground truth in the data generation process in nature (e.g., cyclic).
In addition to that, the number of observations is limited. 
The two challenges mentioned previously contribute to the difficulties faced by traditional causal discovery algorithms in producing accurate predictions.
For example, NHD being 0.33 (NOTEARS and CGNN) is equivalent to NHD of an empty graph. Indeed, CGNN actually outputs an empty adjacency matrix.

\begin{table*}[ht]
\scriptsize\centering
\begin{tabular}{@{}l|lllllll@{}}
\toprule
Method & NHD ($\downarrow$) & NHD Ratio ($\downarrow$) & SHD ($\downarrow$) & No. Edge & FDR ($\downarrow$) & FPR ($\downarrow$) & TPR ($\uparrow$) \\
\midrule
GPT-4 &  0.21 &  0.34 & 19& 43 & 0.30 & 0.13 & 0.62 \\
\midrule
NOTEARS &0.33  & 0.77 & 28 & 14  & 0.50 & 0.07 & 0.14 \\
\phantom{..}w/ random prior &0.44 
$({\color{blue}\blacktriangle} 0.11)$  & 0.60 $({\color{red}\blacktriangledown} 0.17)$ &  37 $({\color{blue}\blacktriangle} 5)$ & 56
& 0.63 
$({\color{blue}\blacktriangle} 0.13)$ & 0.37 $({\color{blue}\blacktriangle} 0.30)$ & 0.43 $({\color{red}\blacktriangle} 0.29)$ \\
\phantom{..}w/ GPT-4 prior &0.24 
$({\color{red}\blacktriangledown} 0.09)$  & 0.46 $({\color{red}\blacktriangledown} 0.31)$ &  17 $({\color{red}\blacktriangledown} 11)$ & 27 
& 0.25 $({\color{red}\blacktriangledown} 0.25)$ & 0.07 & 0.41 $({\color{red}\blacktriangle} 0.27)$ \\
\midrule
CGNN(*) & 0.33 & 0.33 & 48& \phantom{0}0 & - & - & -\\
\phantom{..}w/ random prior &0.42 
$({\color{blue}\blacktriangle} 0.09)$  & 0.66 $({\color{red}\blacktriangledown} 0.33)$ &  39 $({\color{red}\blacktriangledown} 9)$ & 43
& 0.64  & 0.28 & 0.31 \\
\phantom{..}w/ GPT-4 prior &0.22 
$({\color{red}\blacktriangledown} 0.11)$  & 0.39 $({\color{blue}\blacktriangle} 0.06)$ &  19 $({\color{red}\blacktriangledown} 29)$ & 35 & 0.28  & 0.10  & 0.52 \\
\midrule
DAG-GNN &  0.31 &  0.76 & 27& 12 &  0.41 &  0.05 &  0.14 \\
\phantom{..}w/ random prior &0.41 $({\color{blue}\blacktriangle} 0.10)$  & 0.64 $({\color{red}\blacktriangledown} 0.12)$  &  37 $({\color{blue}\blacktriangle} 10)$ & 44 & 0.62 $({\color{blue}\blacktriangle} 0.21)$ & 0.29 $({\color{blue}\blacktriangle} 0.24)$ & 0.33 $({\color{red}\blacktriangle} 0.19)$ \\
\phantom{..}w/ GPT-4 prior &  0.20 $({\color{red}\blacktriangledown} 0.11)$ &  0.33 $({\color{red}\blacktriangledown} 0.43)$ &  18 $({\color{red}\blacktriangledown} 9)$ & 42 & 0.28 $({\color{red}\blacktriangledown} 0.13)$ & 0.12 $({\color{blue}\blacktriangle} 0.07)$ & 0.62 $({\color{red}\blacktriangle} 0.48)$\\
\bottomrule
\end{tabular}
\caption{
Performances of NOTEARS, CGNN, and DAG-GNN on the \textbf{Arctic Sea Ice dataset}.
For each algorithm, with and without GPT-4 prior, and uniform random prior whose number of the edge is the same with GPT-4 prior are investigated.
}
\label{tab:arctic_nhd}
\end{table*}

\begin{figure*}[t]
    \centering
    \includegraphics[width=\textwidth]{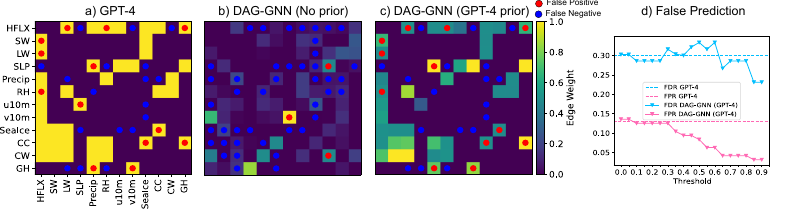}
    \caption{
    Outputs by (a) GPT-4, (b) DAG-GNN, and (c) DAG-GNN with GPT-4 prior. (d) false prediction of DAG-GNN.
    }
    \label{fig:heatmap_daggnn}
\end{figure*}

In contrast, the application of the proposed framework induces overall performance improvement with a big margin compared to causal discovery algorithms.
CGNN's performance generally improves upon integrating the PLM prior, with the exception of the NHD ratio (comparisons using FDR, FPR, and TPR metrics are not feasible since CGNN predicts no edges.) 
This improvement is attributable to the utilization of a well-constructed underlying graph skeleton by PLM. 
The low NHD ratio of CGNN can be attributed to the fact that when there are not any predicted edges, the denominator of the NHD ratio becomes 1 so its NHD ratio is low. 
Interestingly, when incorporating PLM prior knowledge, the FDR decreases by 25\%p for NOTEARS and 13\%p for DAG-GNN, in contrast to FPR. 
This means integrating PLM prior knowledge significantly improves the credibility of model prediction on causal relations. 

Next, when compared to the GPT-4 prior, our methodology exhibited both areas of improvement and instances of slightly reduced performance. 
We illustrate the results for DAG-GNN using heatmaps in \Cref{fig:heatmap_daggnn}.
When compared \Cref{fig:heatmap_daggnn} (b) and (c), DAG-GNN generally showed a slight improvement in performance when using PLM prior compared to GPT-4. 
As shown in \Cref{fig:heatmap_daggnn} d), increasing the threshold results in an increase in true negative, leading to a decrease in the FPR.
While the reduction in predictions leads to a decrease in true positives and, consequently, FDR, the reduction in false positives is more pronounced, contributing to its decrease.
NOTEARS and CGNN exhibited a decline in performance in NHD, NHD ratio, and TPR, but they demonstrated better scores in terms of SHD, FDR, FPR, and TPR, compared with GPT-4.
This result is interpreted as follows: these algorithms predicted a lower number of edges, which consequently reduced the number of false positives; however, this also led to a slight decrease in the number of true positives.


In addition, to highlight the advantages of a high-quality prior, we also contrasted the results with those obtained using a random prior (randomly selected 43 edges, where 43 comes from the number of edges by GPT-4). 
Based on 20 repeated trials, the experimental outcomes consistently aligned with our assumptions, demonstrating higher performance when a better prior was utilized, while predicting similar numbers of edges.

\begin{table*}[ht]
\scriptsize\centering
\begin{tabular}{@{}l|lllllll@{}}
\toprule
Method & NHD ($\downarrow$) & NHD Ratio ($\downarrow$) & SHD ($\downarrow$) & No. Edge & FDR ($\downarrow$) & FPR ($\downarrow$) & TPR ($\uparrow$) \\
\midrule
GPT-4 &  0.19 &  0.53 & 21& 24 & 0.58 & 0.13 & 0.52 \\
\midrule
NOTEARS &  0.19 & 0.67 & 22 & 18  & 0.66 & 0.11 & 0.31 \\
\phantom{..}w/ random prior &  0.27 $({\color{blue}\blacktriangle} 0.09)$ & 0.82 $({\color{blue}\blacktriangle} 0.15)$ & 28 $({\color{blue}\blacktriangle} 6)$ & 21  & 0.83 $({\color{blue}\blacktriangle} 0.17)$ & 0.17 $({\color{blue}\blacktriangle} 0.06)$ & 0.18 $({\color{blue}\blacktriangledown} 0.13)$\\
\phantom{..}w/ GPT-4 prior &  0.18 $({\color{red}\blacktriangledown} 0.01)$ & 0.68 $({\color{blue}\blacktriangle} 0.01)$ & 20 $({\color{red}\blacktriangledown} 2)$ & 13  & 0.61 $({\color{red}\blacktriangledown} 0.05)$ & 0.07 $({\color{red}\blacktriangledown} 0.04)$ & 0.26 $({\color{blue}\blacktriangledown} 0.05)$ \\
\midrule
CGNN & 0.26  & 0.84 & 30 & 19  & 0.84 & 0.15 & 0.15 \\
\phantom{..}w/ random prior &0.29 $({\color{blue}\blacktriangle} 0.03)$  & 0.84 &  31 $({\color{blue}\blacktriangle} 1)$
& 23 
& 0.85 $({\color{blue}\blacktriangle} 0.01)$ & 0.20 $({\color{blue}\blacktriangle} 0.05)$ & 0.17 $({\color{red}\blacktriangle} 0.02)$ \\
\phantom{..}w/ GPT-4 prior &0.14 $({\color{red}\blacktriangledown} 0.12)$  & 0.47 $({\color{red}\blacktriangledown} 0.37)$ &  18 $({\color{red}\blacktriangledown} 12)$ & 19 & 0.47 $({\color{red}\blacktriangledown} 0.37)$ & 0.08 $({\color{red}\blacktriangledown} 0.07)$ & 0.52 $({\color{red}\blacktriangle} 0.37)$ \\
\midrule
DAG-GNN & 0.18  & 0.68 &  19 & 13 & 0.61 &  0.07 &  0.26   \\
\phantom{..}w/ random prior &  0.27 $({\color{blue}\blacktriangle} 0.09)$ & 0.81 $({\color{red}\blacktriangledown} 0.13)$ &  29 $({\color{blue}\blacktriangle} 10)$ & 21 & 0.83 $({\color{red}\blacktriangledown} 0.22)$ & 0.17 $({\color{blue}\blacktriangle} 0.10)$ & 0.20 $({\color{red}\blacktriangle} 0.06)$ \\
\phantom{..}w/ GPT-4 prior &  0.16 $({\color{red}\blacktriangledown} 0.02)$  & 0.47 $({\color{red}\blacktriangledown} 0.21)$ & 19 & 23 & 0.52 $({\color{red}\blacktriangledown} 0.09)$ & 0.11 $({\color{blue}\blacktriangle} 0.04)$ & 0.57 $({\color{red}\blacktriangle} 0.31)$ \\
\bottomrule
\end{tabular}
\caption{
Performances of NOTEARS, CGNN, and DAG-GNN on the \textbf{Sachs dataset}.
For each algorithm, with and without GPT-4 prior, and uniform random prior whose number of the edge is same with GPT-4 prior are investigated. 
}
\label{tab:sachs}
\end{table*}

\paragraph{Sachs}

The performance trends exhibited different behavior across the causal discovery algorithms, as reported in \Cref{tab:sachs}.
The performance of CGNN and DAG-GNN is generally improved when supported by PLM prior, except for the FPR of DAG-GNN.
In the case of NOTEARS, similar results were consistently obtained without normalization of data, even with random priors.
This phenomenon can be attributed that 
NOTEARS is sometimes trapped on local optima on Sachs dataset, even varying the regularization coefficient $\lambda$ of the prior similarity.

In contrast, CGNN exhibited improved performance, because the quality of its vanilla skeleton reports lower performance than PLM prior, indicating a potential for further enhancements.
Compared to PLM, the NHD and SHD metrics showed overall enhancements, although the TPR was somewhat lower because the number of predicted edges is lower than PLM in \Cref{tab:sachs}. 

Overall improvements were observed for DAG-GNN, except for the FPR of DAG-GNN.
The reason vanilla DAG-GNN recorded a lower FPR without the PLM prior is that it predicted causal relations at roughly half the number of our framework.
On the other hand, by increasing edge predictions, our model improved performance with respect to NHD, NHD ratio, SHD, and FDR with true positives.
This underscores the usefulness of our framework, as it not only matched but also surpassed PLM's performance in the envisioned scenarios.
That is, the improvement in FDR and FPR in every algorithm, compared to TPR, resulted in an overall increase in performance, as evidenced by NHD and SHD.



\paragraph{Comprehensive insights}
Based on the comprehensive analysis of both synthetic and real datasets, the employment of a PLM prior mostly outperforms traditional causal discovery methods in both scenarios. 
With respect to the complexity of the data generating process, NOTEARS exhibits high performance for the synthetic dataset of simple linear assumption.
In contrast, DAG-GNN performs well in both linear assumption and non-linear real-world datasets, consistently demonstrating enhancement in overall performance via applying our framework.
In the Arctic Sea Ice dataset, vanilla CGNN fails to predict any edge, but when equipped with adequate PLM prior, it showed overall enhancement in performance metrics.

Concerning the Arctic Sea Ice dataset, although the causal graph is not entirely reliable, we observed that when PLM prior provides a proper prediction for the causal graph, applying our framework surpasses conventional causal discovery algorithms in performance. 
Concurrently, our experiments with  Sachs and synthetic datasets have ascertained that our framework excels even in data-rich scenarios where causal discovery algorithms are not affected by the data scarcity issue. 
In other words, in situations where either PLM stands out, or causal discovery prevails, our framework suggests the potential to outperform both approaches.

\section{Discussion}

Although we have shown promising results of our framework, there exist some possible improvements to our framework by 1) incorporating the uncertainty of PLM's prediction, 2) designing a better prompt, and 3) reorganizing the sequence of operations for PLM-based causal reasoning and the causal discovery algorithm within our framework.
\paragraph{Uncertainty measure for PLM prior} 
As noted in \Cref{introduction}, since PLM stochastically predicts causal relations without data, fully trusting $\mathbf{K}$ may make it hard to find the causal structure.
On this note, it would be informative for later research to explore the appropriate design of uncertainty in PLM-based causal reasoning when integrating prior knowledge $\mathbf{K}$ into causal discovery.
When multiple prior knowledge matrices are available, we suggest defining and incorporating a certainty matrix $\mathbf{C} \in \mathbb{R}^{d \times d}$ where each element represents the certainty of its corresponding edge. 
One way to define $\mathbf{C}$ is based on the element-wise reciprocal of the standard deviation of $\mathbf{K}$,
$
C_{i,j} \coloneqq 
    \frac {\epsilon} {\sqrt{\text{Var}[K_{i,j} ]} + \epsilon} 
$
where $\epsilon$ is a hyperparameter for adjusting certainty when $K_{i,j}$ nears to zero.

\paragraph{Prompt design}
The prompt design of \Cref{fig:causalreasoning-a} and \cite{kichiman2023} is multiple-choice templates deciding the existence and direction of an edge.
However, the prompt designs have no `unknown' option for PLM, resulting 
in over-confident predictions by PLM 
even when PLM has little or no understanding of the variables. 
In addition, the aforementioned prompts only focus on the relationship between two variables, failing to properly capture chain structures and other intricate causal patterns.
The new prompt design that takes these considerations into account 
will bring out qualitative or quantitative improvement of our framework.

\paragraph{Reorganizing the sequence of operations}

Although we conducted causal discovery with the prior knowledge generated by PLM, one can consider inverting the framework structure using the outcomes from causal discovery as prior knowledge for PLM-based causal reasoning. 
In devising an inverted framework, one needs to carefully develop a technique to inject the output of causal discovery algorithm in the form of text into PLM---%
naively translating the resulting graph as a prompt (e.g., listing all the edges) may result in a prohibitively long prompt when the number of variables is large, and it is generally known the longer the prompt is, the harder for PLM to understand the underlying meaning.
One possible limitation of this approach is that the outcome of causal reasoning is an adjacency matrix made of 0 and 1, thus, lacking fine-grained information such as structural coefficients.


\section{Conclusion}
We demonstrated the utilization of PLM in causal discovery algorithms through experiments on both synthetic and real data.
%
We illustrated through a physical commonsense synthetic dataset that PLM-based causal reasoning is prone to false prediction.
To tackle the problem, we proposed a novel framework that incorporates the prior knowledge extracted from PLMs into score-based causal discovery algorithms.
The integration is achieved through graph initialization and regularization, leveraging PLM-based causal reasoning. 
This approach combines the strengths of both worlds: reducing the potential for false predictions of PLMs by applying data-driven structural learning from the causal discovery algorithm and enhancing causal discovery performance by incorporating prior knowledge extracted from PLMs.
In our experiments, we observed that 
our framework could fix the PLM's false causal relations owing to learning available data,
improving several important performance metrics.
We have extended the existing realm of causal discovery through the integration with PLM, thereby unveiling new potentials within causal discovery. 
We expect that our framework will open up new avenues for research and exploration in causal discovery.
For future studies, it needs to modify the limitations of the current prompt to extract prior knowledge of PLMs. 
Moreover, our framework can not address time series data, so we need an extensive framework that can be applied to time series causal discovery.

\bibliographystyle{abbrv}
\bibliography{arxiv}

\begin{thebibliography}{10}

\bibitem{addo2021exploring}
P.~M. Addo, C.~Manibialoa, and F.~McIsaac.
\newblock Exploring nonlinearity on the co2 emissions, economic production and
  energy use nexus: a causal discovery approach.
\newblock {\em Energy Reports}, 7:6196--6204, 2021.

\bibitem{anil2023palm}
R.~Anil et~al.
\newblock Palm 2 technical report, 2023.

\bibitem{bisk2019piqa}
Y.~Bisk, R.~Zellers, R.~L. Bras, J.~Gao, and Y.~Choi.
\newblock Piqa: Reasoning about physical commonsense in natural language.
\newblock In {\em AAAI Conference on Artificial Intelligence}, 2019.

\bibitem{Borboudakis2011ACA}
G.~Borboudakis, S.~Triantafillou, V.~Lagani, and I.~Tsamardinos.
\newblock A constraint-based approach to incorporate prior knowledge in causal
  models.
\newblock In {\em The European Symposium on Artificial Neural Networks}, 2011.

\bibitem{Borboudakis2012IncorporatingCP}
G.~Borboudakis and I.~Tsamardinos.
\newblock Incorporating causal prior knowledge as path-constraints in bayesian
  networks and maximal ancestral graphs.
\newblock In {\em International Conference on Machine Learning}, 2012.

\bibitem{Chen2021EvaluatingLL}
M.~Chen et~al.
\newblock Evaluating large language models trained on code.
\newblock {\em ArXiv}, abs/2107.03374, 2021.

\bibitem{Chickering2003OptimalSI}
D.~M. Chickering.
\newblock Optimal structure identification with greedy search.
\newblock {\em J. Mach. Learn. Res.}, 3:507--554, 2003.

\bibitem{choi2022lmpriors}
K.~Choi, C.~Cundy, S.~Srivastava, and S.~Ermon.
\newblock Lmpriors: Pre-trained language models as task-specific priors.
\newblock {\em ArXiv}, abs/2210.12530, 2022.

\bibitem{de2004discovery}
A.~De~La~Fuente, N.~Bing, I.~Hoeschele, and P.~Mendes.
\newblock Discovery of meaningful associations in genomic data using partial
  correlation coefficients.
\newblock {\em Bioinformatics}, 20(18):3565--3574, 2004.

\bibitem{GlymourZhangSpirtes2019_reviewb}
C.~Glymour, K.~Zhang, and P.~Spirtes.
\newblock Review of {{Causal Discovery Methods Based}} on {{Graphical Models}}.
\newblock {\em Frontiers in Genetics}, 10, 2019.

\bibitem{goudet2018learning}
O.~Goudet, D.~Kalainathan, P.~Caillou, I.~Guyon, D.~Lopez-Paz, and M.~Sebag.
\newblock Learning functional causal models with generative neural networks.
\newblock {\em Explainable and interpretable models in computer vision and
  machine learning}, pages 39--80, 2018.

\bibitem{huang2021arctic}
Y.~Huang, M.~Kleindessner, A.~Munishkin, D.~Varshney, P.~Guo, and J.~Wang.
\newblock Benchmarking of data-driven causality discovery approaches in the
  interactions of arctic sea ice and atmosphere.
\newblock {\em Frontiers in Big Data}, 2021.

\bibitem{kalainathan1903causal}
D.~Kalainathan and O.~Goudet.
\newblock Causal discovery toolbox: Uncover causal relationships in python,
  2019.
\newblock {\em URL https://arxiv. org/abs}, 1903.

\bibitem{Kalainathan2018SAMSA}
D.~Kalainathan, O.~Goudet, I.~M. Guyon, D.~Lopez-Paz, and M.~Sebag.
\newblock Sam: Structural agnostic model, causal discovery and penalized
  adversarial learning.
\newblock {\em arXiv: Machine Learning}, 2018.

\bibitem{kingma2013vae}
D.~P. Kingma and M.~Welling.
\newblock Auto-encoding variational bayes.
\newblock {\em CoRR}, abs/1312.6114, 2013.

\bibitem{kichiman2023}
E.~Kıcıman, R.~Ness, A.~Sharma, and C.~Tan.
\newblock Causal reasoning and large language models: Opening a new frontier
  for causality, 2023.

\bibitem{li2009controlling}
J.~Li and Z.~J. Wang.
\newblock Controlling the false discovery rate of the association/causality
  structure learned with the pc algorithm.
\newblock {\em Journal of Machine Learning Research}, 10(2), 2009.

\bibitem{openai2023gpt4}
OpenAI.
\newblock Gpt-4 technical report, 2023.

\bibitem{ramsey2018fask}
J.~Ramsey and B.~Andrews.
\newblock Fask with interventional knowledge recovers edges from the sachs
  model.
\newblock {\em arXiv preprint arXiv:1805.03108}, 2018.

\bibitem{sachs2005causal}
K.~Sachs, O.~Perez, D.~Pe'er, D.~A. Lauffenburger, and G.~P. Nolan.
\newblock Causal protein-signaling networks derived from multiparameter
  single-cell data.
\newblock {\em Science}, 308(5721):523--529, 2005.

\bibitem{Sap2019ATOMICAA}
M.~Sap, R.~L. Bras, E.~Allaway, C.~Bhagavatula, N.~Lourie, H.~Rashkin, B.~Roof,
  N.~A. Smith, and Y.~Choi.
\newblock Atomic: An atlas of machine commonsense for if-then reasoning.
\newblock {\em ArXiv}, abs/1811.00146, 2019.

\bibitem{sinha2021perturbing}
S.~Sinha, H.~Chen, A.~Sekhon, Y.~Ji, and Y.~Qi.
\newblock Perturbing inputs for fragile interpretations in deep natural
  language processing.
\newblock In {\em Proceedings of the Fourth BlackboxNLP Workshop on Analyzing
  and Interpreting Neural Networks for NLP}, pages 420--434, Punta Cana,
  Dominican Republic, Nov. 2021. Association for Computational Linguistics.

\bibitem{spirtes2000causation}
P.~Spirtes, C.~N. Glymour, and R.~Scheines.
\newblock {\em Causation, prediction, and search}.
\newblock MIT press, 2000.

\bibitem{spirtes2016causal}
P.~Spirtes and K.~Zhang.
\newblock Causal discovery and inference: concepts and recent methodological
  advances.
\newblock {\em Applied Informatics}, 3(1):3, 2016.

\bibitem{Suzgun2022ChallengingBT}
M.~Suzgun, N.~Scales, N.~Scharli, S.~Gehrmann, Y.~Tay, H.~W. Chung,
  A.~Chowdhery, Q.~V. Le, E.~H. hsin Chi, D.~Zhou, and J.~Wei.
\newblock Challenging big-bench tasks and whether chain-of-thought can solve
  them.
\newblock In {\em Annual Meeting of the Association for Computational
  Linguistics}, 2022.

\bibitem{Thoppilan2022LaMDALM}
R.~Thoppilan et~al.
\newblock Lamda: Language models for dialog applications.
\newblock {\em ArXiv}, abs/2201.08239, 2022.

\bibitem{touvron2023llama}
H.~Touvron et~al.
\newblock Llama 2: Open foundation and fine-tuned chat models, 2023.

\bibitem{tsamardinos2006max}
I.~Tsamardinos, L.~E. Brown, and C.~F. Aliferis.
\newblock The max-min hill-climbing bayesian network structure learning
  algorithm.
\newblock {\em Machine learning}, 65:31--78, 2006.

\bibitem{Wei2022EmergentAO}
J.~Wei, Y.~Tay, R.~Bommasani, C.~Raffel, B.~Zoph, S.~Borgeaud, D.~Yogatama,
  M.~Bosma, D.~Zhou, D.~Metzler, E.~H. hsin Chi, T.~Hashimoto, O.~Vinyals,
  P.~Liang, J.~Dean, and W.~Fedus.
\newblock Emergent abilities of large language models.
\newblock {\em Trans. Mach. Learn. Res.}, 2022, 2022.

\bibitem{wei2023chainofthought}
J.~Wei, X.~Wang, D.~Schuurmans, M.~Bosma, B.~Ichter, F.~Xia, E.~Chi, Q.~Le, and
  D.~Zhou.
\newblock Chain-of-thought prompting elicits reasoning in large language
  models, 2023.

\bibitem{yu2019daggnn}
Y.~Yu, J.~Chen, T.~Gao, and M.~Yu.
\newblock Dag-gnn: Dag structure learning with graph neural networks.
\newblock In {\em International Conference on Machine Learning}, 2019.

\bibitem{zheng2018notears}
X.~Zheng, B.~Aragam, P.~K. Ravikumar, and E.~P. Xing.
\newblock Dags with no tears: Continuous optimization for structure learning.
\newblock In S.~Bengio, H.~Wallach, H.~Larochelle, K.~Grauman, N.~Cesa-Bianchi,
  and R.~Garnett, editors, {\em Advances in Neural Information Processing
  Systems}, volume~31. Curran Associates, Inc., 2018.

\end{thebibliography}

\clearpage\newpage

\appendix


\section{Experimental Details}
\label{supple:experimental setup}


\subsection{Metrics}
\label{supple:Metrics}
SHD is a sum of the number of missing edges (false negative), extra edges (false positive), and reversed edges \cite{tsamardinos2006max}.
NHD is a metric that normalizes the Hamming distance, representing the number of differing edges, dividing the Hamming distance by the matrix size. It yields values between 0 and 1, with lower values indicating greater similarity to the causal graph.
NHD ratio is that NHD divided by baseline NHD which means that the worst case for some specific number of edges so that we can figure out the estimated adjacency matrix how much improved than the worst case.
FDR, FPR, and TPR are derived from the four outcomes of the confusion matrix: False Positive, False Negative, True Positive, and True Negative and these metrics collectively evaluate the errors in classification:
\begin{align*}
\text{FDR} &= \frac{\text{FP}}{\text{FP} + \text{TP}}, \quad \text{FPR} = \frac{\text{FP}}{\text{FP} + \text{TN}}, \quad \text{TPR} = \frac{\text{TP}}{\text{TP} + \text{FN}}.
\end{align*}

\subsection{Setup}
\label{supple:setup}
The baseline code was referenced from \cite{kalainathan1903causal, yu2019daggnn}, CausalNex (https://github.com/quantumblacklabs/causalnex).
The experimental setup, such as hyperparameter and model architecture, is as follows:
First, we detail the hyperparameter---%
$t$, $\lambda_{sim}$, thresholds 
in \Cref{tab:hyperparameter_arctic,tab:hyperparameter_sachs}.
$\lambda_{init}$ is the scaling factor for graph initialization and $\lambda_{sim}$ is that for prior similarity regularization.
As we mentioned in the Experimental setup of \Cref{sec:experiments}, hyperparameters of baseline were tuned to reproduce baseline experiments, and that of our experiments were selected for adequate demonstration of our framework.
For NOTEARS, the activation function used in \Cref{eq:sim} is a sigmoid function.
The hyperparameter for the sigmoid activation function $t$ is 10. 
For DAG-GNN, we used clamping with $[0,1]$ instead of the sigmoid function.
Second, the model architecture and other setups are as follows.
For DAG-GNN, we used the Adam optimizer and two layers each for the encoder and the decoder. 
We allocated 64 hidden nodes in each layer for the Arctic Sea Ice model and 128 hidden nodes in each layer for the Sachs model, with a uniform batch size set at 100 for DAG-GNN.
Though the experiments are feasible on CPUs, our experiments were primarily conducted using NVIDIA RTX A6000 and Tesla V100-SXM2-32GB GPUs.
\begin{table}[t]
\footnotesize\centering
\begin{tabular}{l|llll}
\toprule
Method &  Prior & $\lambda_{init}$ & $\lambda_{sim}$ & threshold  \\
\midrule
NOTEARS &  GPT-4 & 1 & 0.7 & 0.1  \\
CGNN & GPT-4 & 1 & - & 0.99  \\
DAG-GNN & GPT-4 & 0.3 & 0.7 & 0.3  \\
NOTEARS &  None & - & - & 0.13  \\
CGNN &  None & - & - & -  \\
DAG-GNN &  None & - & - & 0.3  \\
\bottomrule
\end{tabular}

\caption{
Hyperparameters of the Arctic Sea Ice for each algorithm and each dataset
}
\label{tab:hyperparameter_arctic}
\end{table}

\begin{table}[t]
\footnotesize\centering
\begin{tabular}{l|llll}
\toprule
Method &  Prior & $\lambda_{init}$ & $\lambda_{sim}$ & threshold  \\
\midrule
NOTEARS &  GPT-4 & 1 & 0.05 & 0.57  \\
CGNN & GPT-4 & 1 & - & 0.65  \\
DAG-GNN & GPT-4 & 0.7 & 0.7 & 0.6  \\
NOTEARS &  None & - & - & 0.16  \\
CGNN &  None & - & - & -  \\
DAG-GNN &  None & - & - & 0.3  \\
\bottomrule
\end{tabular}

\caption{
Hyperparameters of the Sachs for each algorithm and each dataset
}
\label{tab:hyperparameter_sachs}
\end{table}

\section{Physical Commonsense-Based Synthetic Dataset}

\label{supple:generation}
\subsection{Construction of Causal Graph Based on Physical Commonsense}
\label{supple:physical}
In this section, we explain how to construct a physical commonsense-based synthetic dataset for evaluating causal discovery algorithms and the causal reasoning ability of PLM.
To evaluate the reasoning ability of PLM, we chose to construct a knowledge base of a specific domain.
Because causal reasoning focuses on strict logical relations between variables, the annotated content based on the selected domain should contain clear ground truth.
For this reason, domains dealing with not exactly decisive problems, such as social or cultural domains, are unsuitable, so we decided to construct a knowledge base on physics.

We utilized the PIQA \cite{bisk2019piqa} to select the proper physical event where causal relationships hold.
We removed text that is ambiguous or described too specifically from our knowledge base.
We selectively annotated entities that describe phase transition.
Phase transition refers to phenomena where a matter's phase, such as solid, liquid, or gas, transit to another phase.
For example, the increase in `surface air temperature' causes a change in the evaporation rate of water, transferring the object from the liquid phase to the gas phase.
Using this strategy to annotate the PIQA dataset, we focused on the physics of water evaporation, and we found entities involved in the evaporation process.
With the annotated entities, we gathered nodes of a causal graph whose nodes are entities involved in the phase transition, and human annotators evaluate the causal relationships among the nodes.




\subsection{Datasets Generation Given Synthetic Graph} 
To generate a synthetic dataset based on a physical commonsense-based causal graph, we selected seven nodes that represent the evaporation of water such that collected nodes and edges satisfy the DAG constraint.
Given the causal graph composed of seven nodes, we added subgraphs of five and three nodes from the predefined graph by removing the proper node for subgraph construction.
Removing nodes, we add additional edges from ancestor to descendent whenever the removed node connects the ancestor and descendent so that the chain relation holds.
Using the constructed 3, 5, and 7 nodes graphs, we assumed a linear Structural Equation Model between variables and Gaussian noise of $\epsilon \sim \mathcal{N}(0, 0.5)$ within a given causal graph and generate 5000 data points.

\section{Other Experimental Results}

\begin{figure}[ht]
    \centering
    \includegraphics[width=.8\columnwidth]{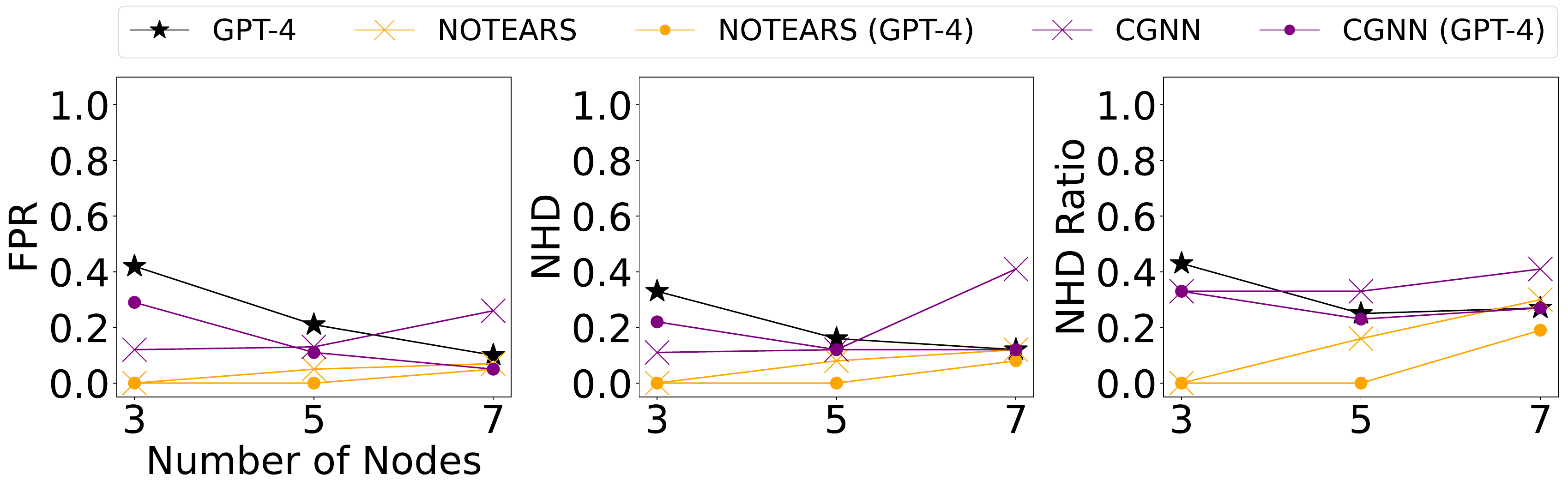}
    \caption{
    FPR, NHD, NHD Ratio of comparison on the physical knowledge-based synthetic datasets.
    }
    \label{fig:synthetic results2}
\end{figure}

\begin{figure*}[t]
    \centering
    \includegraphics[width=0.7\textwidth]{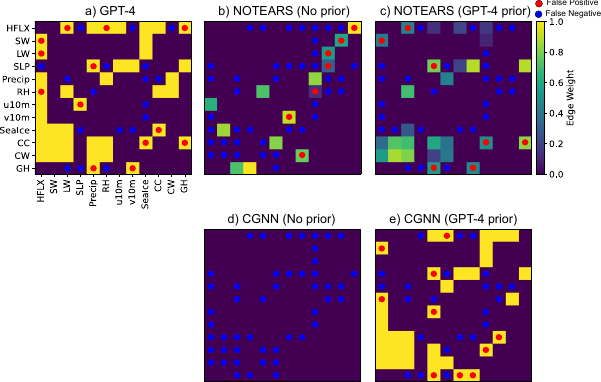}
    \caption{
    Heatmaps in Arctic Sea Ice dataset by 
    a) GPT-4, 
    b) NOTEARS, and c) NOTEARS with GPT-4 prior, 
    d) CGNN, e) CGNN with GPT-4 prior.
    }
    \label{fig:heatmap_arctic_supple}
\end{figure*}

\begin{figure*}[t]
    \centering
    \includegraphics[width=0.7\textwidth]{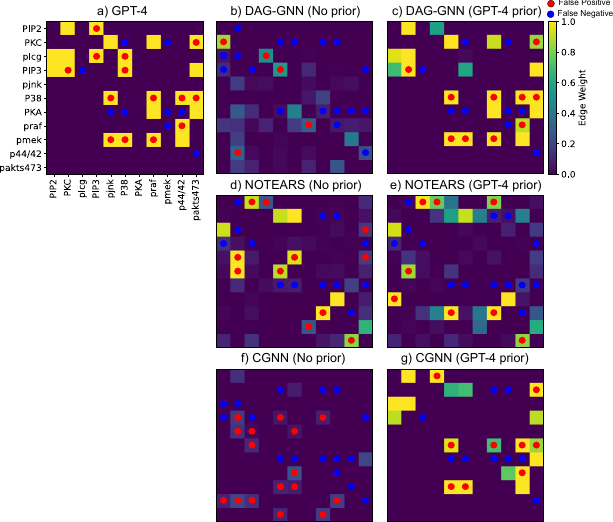}
    \caption{
    Heatmaps in Sachs dataset by 
    a) GPT-4, 
    b) DAG-GNN, and c) DAG-GNN with GPT-4 prior, 
    d) NOTEARS, and e) NOTEARS with GPT-4 prior, 
    f) CGNN, g) CGNN with GPT-4 prior.
    }
    \label{fig:heatmap_sachs_supple}
\end{figure*}

We also conducted other experimental results and figures. 
\Cref{fig:synthetic results2} shows FPR, NHD, NHD ratio of NOTEARS and CGNN on physical knowledge-based synthetic datasets whose sizes are 3, 5, and 7 nodes.
\Cref{fig:heatmap_arctic_supple} and \Cref{fig:heatmap_sachs_supple} are heatmaps for NOTEARS and CGNN for each dataset.



\end{document}